\documentclass[runningheads]{llncs}

\usepackage[mobile]{eccv}

\usepackage{eccvabbrv}
\usepackage{graphicx}
\usepackage{booktabs}
\usepackage{multirow}
\usepackage[utf8]{inputenc}
\usepackage{fourier} 
\usepackage{array}
\usepackage{makecell}
\usepackage{enumerate}  
\usepackage[usestackEOL]{stackengine}
\usepackage[accsupp]{axessibility}  
\usepackage{subcaption}

\usepackage[pagebackref,breaklinks,colorlinks,citecolor=eccvblue]{hyperref}

\usepackage{orcidlink}

\begin{document}

\title{STMPL: Human Soft-Tissue Simulation} 


\author{Anton Agafonov \and
Lihi Zelnik-Manor }

\authorrunning{A.~Agafonov et al.}

\institute{Technion - Israel Institute of Technology}

\maketitle

\begin{abstract}

In various applications, such as virtual reality and gaming, simulating the deformation of soft tissues in the human body during interactions with external objects is essential. Traditionally, Finite Element Methods (FEM) have been employed for this purpose, but they tend to be slow and resource-intensive. In this paper, we propose a unified representation of human body shape and soft tissue with a data-driven simulator of non-rigid deformations. This approach enables rapid simulation of realistic interactions.

Our method builds upon the SMPL model, which generates human body shapes considering rigid transformations. We extend SMPL by incorporating a soft tissue layer and an intuitive representation of external forces applied to the body during object interactions. Specifically, we mapped the 3D body shape and soft tissue and applied external forces to 2D UV maps. Leveraging a UNET architecture designed for 2D data, our approach achieves high-accuracy inference in real time. Our experiment shows that our method achieves plausible deformation of the soft tissue layer, even for unseen scenarios.

\url{https://antonagafonov.github.io/STMPL/}

\end{abstract}

\section{Introduction}

Accurately simulating the soft-tissue deformation of the human body during interactions with external objects or other individuals is crucial for many applications. With the growing accessibility of virtual and augmented reality technologies, there is an increasing demand for high-fidelity simulations in various domains, for example, virtual medical diagnostics~\cite{Nguyen_Ho_Ba_Tho_Dao_2020}, surgical planning~\cite{10.1115/1.4055835},\cite{PhysGNN},\cite{Human_Touch}, medical training, and virtual try-on experiences in e-commerce \cite{halimi2023physgraph},\cite{grigorev2023hood}.

The go-to approach for creating realistic soft tissue deformations is physics-based Finite Element Methods (FEM) ~\cite{polyfem}. FEM offers the benefit of generating authentic deformations for almost any given object with different properties and geometry. However, achieving accurate simulations necessitates high-resolution meshes, leading to increased computational load that limits real-time rendering.
Faster solutions for simulating human body deformations~\cite{Meekyoung:siggraph},\cite{https://doi.org/10.1111/cgf.13913} propose a hybrid approach that first generates the rigid body transformation using extensions of the Skinned Multi-Person Linear (SMPL) model~\cite{SMPL:2015}, and then the soft tissue deformation is simulated using FEM. The hybrid approach is faster than the fully FEM-based one but is still not a real-time solution.

\begin{figure}[tbh]
    \centering
    \begin{subfigure}[t]{0.4\linewidth}
        \centering
        \includegraphics[width=\linewidth]{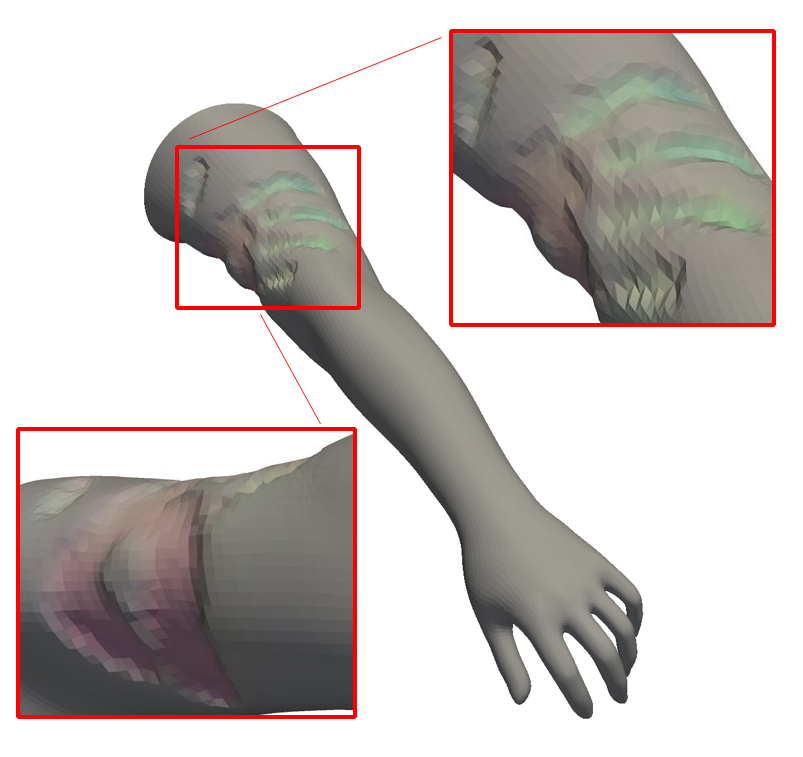}
        \label{fig:1a}
    \end{subfigure}
    \hfill 
    \begin{subfigure}[t]{0.55\linewidth}
        \centering
        \includegraphics[width=0.8\linewidth]{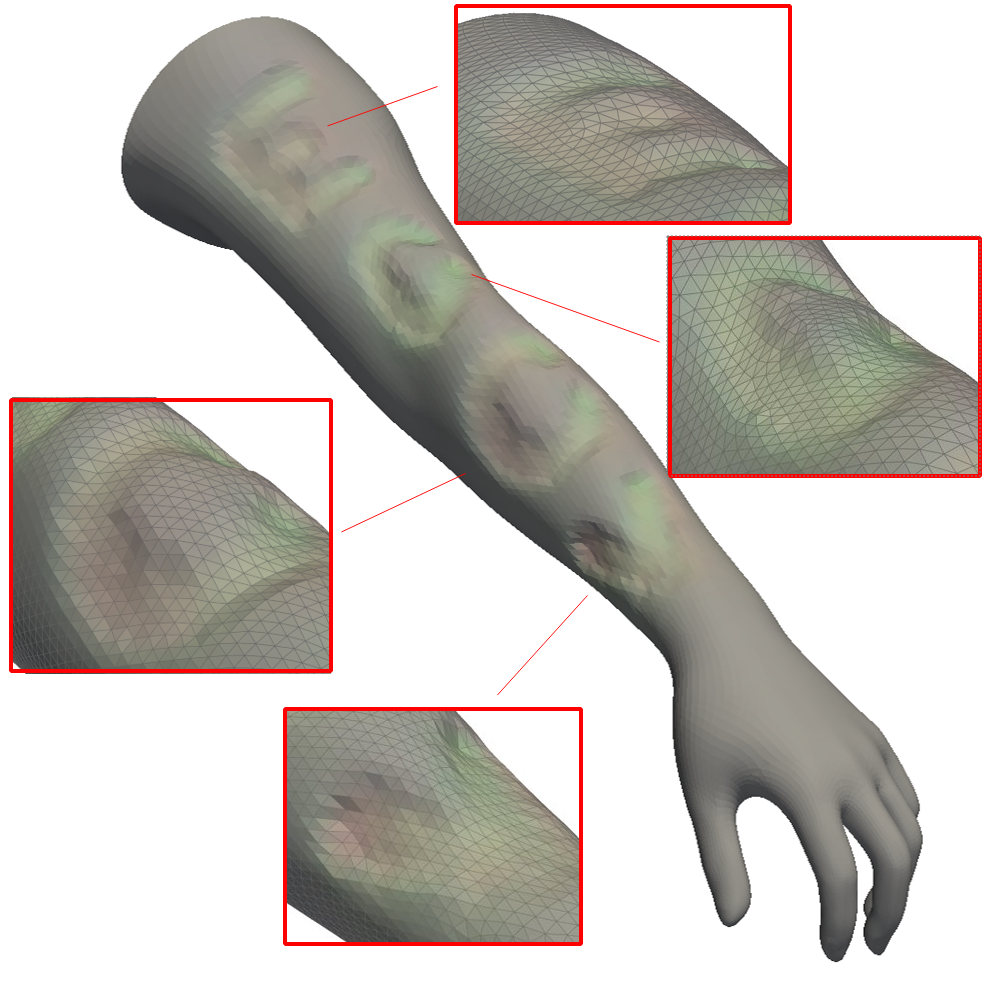}  
        \label{fig:1b}
\end{subfigure}
s


     \caption{\textbf{STMPL. } STMPL's ability to simulate the soft tissue deformation of human bodies in real time is demonstrated through interactions with external interactions of varying shapes. We illustrate the capabilities of the STMPL model when simulating contact with grasping hands and 'ECCV' block letters, showcasing the detailed and realistic deformation such interactions produce.}
     

    \label{fig:teaser_figure}
\end{figure}

In this paper, we propose a learning-based approach that does not require FEM simulation during runtime. To this end, we present the Soft-Tissue-Multi-Person-Linear model (STMPL), which adds a soft tissue layer to the SMPL representation of the human body and is designed to simulate its deformations. 
The STMPL representation provides an efficient and intuitive way to simulate interactions with the human body. 
There are two key ideas behind STMPL: the first is to model the soft tissue as a thick 3D layer of varying thickness, and the second is to map the 3D body mesh, as well as the 3D soft tissue, onto 2D using already existing SMPL, UV parametrization.
This 2D mapping simplifies the representation of the soft tissue layer, as well as the representation of the interacting forces. In addition, it allows us to utilize a technique based on standard UNET for simulating the deformation caused by the interaction with external forces.

Our goal is to enable simulation of interaction with objects of any shape. The straightforward approach to address that would be to build a large training data set, where the ground truth consists of a FEM-based simulation of interactions between numerous human bodies with numerous objects. However, since FEM-based simulation is computationally heavy, we opt to refrain from such data- and resource-hungry approaches. 
Instead, we assert that interaction with complex objects can be learned by combining interactions with simple objects. To put this to the test, we chose to limit the training to interactions with disks and ellipses while generalizing at test time to interactions with complex shapes, such as a grasping hand. 

We demonstrate empirically that the proposed approach works well in practice and that the trained STMPL model generalizes well to a wide range of complex, realistic, and intricate interactions.
Our inference results are comparable to the accuracy of FEM-generated data. Furthermore, the model generalizes well to unseen shapes of interacting objects and soft-tissue thickness variations that were not used during training.
STMPL inference time 0.002 $\pm$ 0.0003 seconds per simulation, compared to 84 $\pm$ 53.4 seconds required for FEM. We intend to make the STMPL dataset used for training and our code public.

\section{Related Work}

Parametric human body models have gained significant attention in computer vision and computer graphics due to their versatility in various applications, such as pose estimation \cite{tripathi20233d}, 3D digital human reconstruction \cite{chen2023gmnerf}, and virtual try-on  \cite{grigorev2023hood},\cite{lazova2019360degree}. A seminal contribution in this domain is the SMPL \cite{SMPL:2015}, which provides a compact and expressive representation of the human body. SMPL captures body shape and pose variations using a linear blend skinning technique and has been widely adopted for its efficiency and accuracy. Since its inception, several extensions and improvements have emerged in the literature, such \cite{MANO:SIGGRAPHASIA:2017}, \cite{STAR:2020}.

Some approaches~\cite{Data_Driven_Physics_for_Human_Soft_Tissue_Animation},~\cite{Modeling_and_Estimation_of_Nonlinear_Skin_Mechanics_for_Animated_Avatars} integrate physics-based and data-driven based modeling, treating the rigid and soft layers separately. Both methods deal with soft tissue deformation caused by dynamic movements of the body and do not deal with interactions with external forces. 
\cite{prwwflhds18} introduces a phenomenological model, the "sliding thick skin" model, that offers a comprehensive approach encompassing the measurement, modeling, parameter estimation, and simulation of soft tissue properties through the utilization of FEM. Their approach excels in accurately modeling and simulating deformations for specific individuals; however, it achieves this by focusing on a particular human body shape and hence is slow to extend to other human body shapes. In contrast, the STMPL method could be easily adapted to different body shapes. 

Simulation methods that do not rely on FEM have also been proposed. \cite{Karami_2023} and \cite{Simulation_of_hyperelastic_materials_in_real_time_using_deep_learning} suggest learning-based techniques to simulate viscoelastic and hyperelastic materials behavior in real-time, bypassing the computational complexity of FEM.
\cite{Image-based_deep_learning_of_finite_element_simulations} focuses on image-based training of deep learning models to learn finite element simulations directly from data, offering a novel paradigm for deformation simulation of a liver.
 In \cite{Thuerey_2020}, an investigation based on the UNET \cite{U-Net} architecture assesses the precision of deep learning models in predicting Reynolds-Averaged Navier-Stokes solutions. 
An important line of work simulates non-rigid deformations using Graph Neural Networks (GNNs), as demonstrated in \cite{PhysGNN}. This method currently necessitates tailored training for specific objects and cannot encompass the entire range of potential body types. In contrast, our approach focuses on acquiring deformation knowledge that spans a broad spectrum of SMPL model parameterizations.

\section{Method}
\label{method}

\begin{figure}[tbh]
    \centering
    \includegraphics[width=\linewidth]{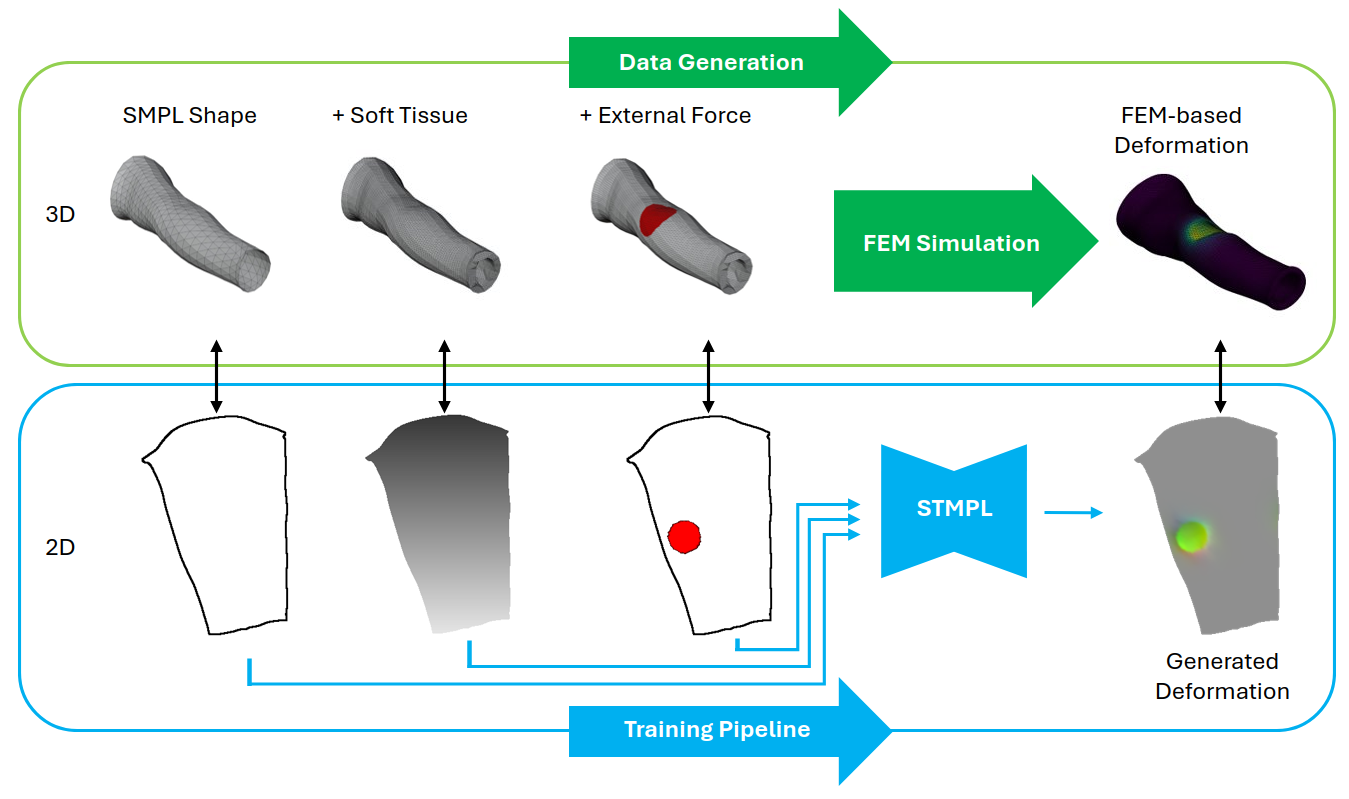}
    \caption{
    \textbf{Method Overview.} STMPL is based on a mapping between 3D (top) to 2D (bottom) of the body shape, the soft tissue layer, and the interacting external force. 
    The generation of training data is conducted utilizing a 3D FEM simulator. The training process (bottom) is executed in a 2D framework, where input parameters such as body shape, soft tissue thickness, and external force are provided, while the FEM-derived deformation, projected onto a 2D space, serves as the ground truth for the output.
        }

    \label{fig:method_overview}
\end{figure}
\begin{figure}[tbh]
    \centering
    \includegraphics[width=\linewidth]{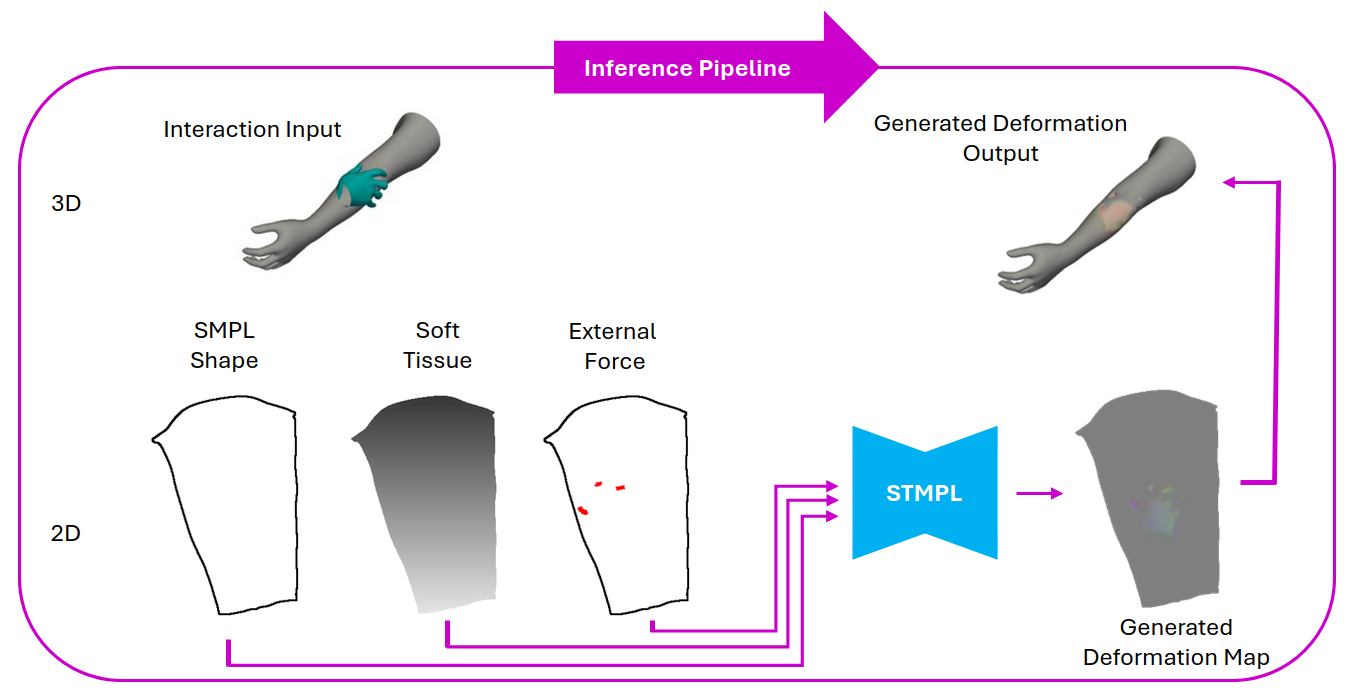}
    \caption{
    \textbf{Inference Pipeline.}
    Inference is performed in 2D and takes as input interactions with complex shapes with variability of soft tissue thickness. The generated 2D deformation is mapped back into 3D to obtain the final deformed mesh.
    }
    \label{fig:testing_overview}
\end{figure}

The underlying infrastructure of our approach is the introduction of the STMPL representation, visualized in Fig. \ref{fig:method_overview}, of human-body interaction, which consists of three 3D components: (i) the human body shape generated by SMPL~\cite{SMPL:2015} (Sec. \ref{SMPL}), (ii) the soft tissue layer modeled as the thickness at each point on the body (Sec. \ref{soft_tissue_layer}), and (iii) the external force applied to the body by the interacting object (Sec. \ref{explain_bc_creation}). 
As shown in Fig. ~\ref{fig:method_overview} (top), data generation of soft deformations resulting from applying the external force to the body is performed in 3D using FEM (Sec. \ref{training_data_generation}). The 3D data is mapped onto 2D maps, which are fed into STMPL simulator for training, as illustrated in Fig. ~\ref{fig:method_overview} (bottom).
Inference is performed in 2D as well, as illustrated in Fig. ~\ref{fig:testing_overview}, and then the generated 2D deformation is mapped back into 3D to obtain the final deformed mesh (Sec. \ref{3D_to_2D}). 



\begin{figure}[t]
    \centering
    \includegraphics[width=12cm]{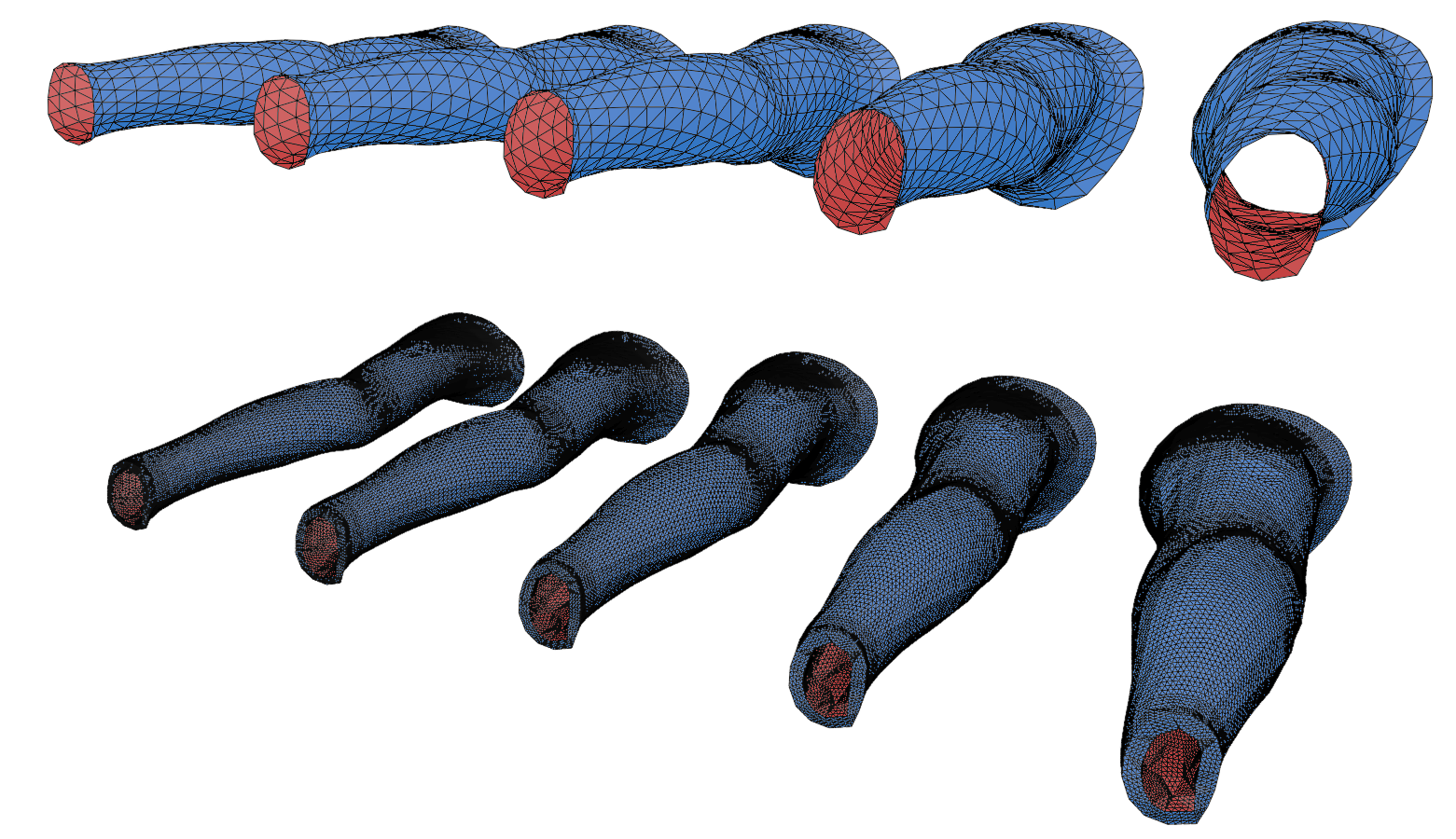}
    \caption{
    \textbf{Soft Layer Generation.}
 Top: five arms generated by the SMPL model with \\ $\beta_{2}$ = [-2,-1,0,1,2]. Bottom: five arms with a range of thickness  [0.5,0.625,0.75, 0.875, 1.0] cm.}
    \label{fig:elastic-body}
\end{figure}

\subsection{Body Surface Model Based on SMPL and UV Parameterization} 
\label{SMPL}
Our method relies on SMPL \cite{SMPL:2015}, a parametric vertex-based model, which represents a wide variety of posed human body surfaces, with $N = 6890$ vertices $V$ and $K=24$ skeletal joints $J$. Vertex positions are defined using two sets of parameters: 1) \textit{pose} $\theta$, $|\theta|=75$ representing the translation and rotation of skeletal joints and 2) \textit{shape} $\beta$, $|\beta|=10$ representing aspects such as height, slenderness or muscularity. SMPL creates a shape-based surface $S(\beta,\theta)$ for a given $\theta $ and $\beta$. For a comprehensive understanding and detailed exposition of SMPL, we refer the reader to~\cite{SMPL:2015}. 

The SMPL model further offers a 2D representation corresponding to the 3D model, a.k.a UV parameterization. It has been typically used to map 2D textures onto the 3D human body model. We propose utilizing the UV parameterization for a different purpose -- to translate the 3D shape onto a 2D UV map with three layers corresponding to the [$x_i$,$y_i$,$z_i$] coordinates of the vertices $V_i$. This allows to further use a UV map to represent the soft deformation a body undergoes via the change in position of each vertex $\delta V_i = [\delta x_i, \delta y_i, \delta z_i]$. 

\subsection{Soft Tissue Representation}
\label{soft_tissue_layer}
The soft tissue is generated as a volumetric layer with varying thicknesses on top of the skinned SMPL model. Inspired by prior works, which focused on dynamic deformation prediction ~\cite{Human_Touch}, \cite{Meekyoung:siggraph},\cite{Modeling_and_Estimation_of_Nonlinear_Skin_Mechanics_for_Animated_Avatars}, we consider the volumetric soft-tissue as an offset surface from the SMPL surface.
The 3D soft tissue layer can be mapped onto its corresponding UV map in a similar fashion to the parametrization of the SMPL model, such that for each vertex $V_i$, there is an appropriate thickness value on the soft tissue UV map.

In our implementation, we represent the soft layer as a constant offset from the skin with fixed material properties. 
During data generation, we wanted to capture a variety of soft tissue thickness values.
To achieve this, we model different soft tissue thicknesses by altering the parameter $\beta$, recognized for its strong correlation with slenderness. Thus, thick bodies implied a thick soft layer, and vice versa.
While this heuristic is oversimplified, it was sufficient to demonstrate our approach's capabilities and can be extended to more complex models in the future.
Specifically, soft-tissue thickness diversity was achieved by setting $\beta$ = [0, $t$, 0, 0, 0, 0, 0, 0, 0, 0], with $t$ = $[2.0, 1.0, 0.0, -1.0, -2.0]$, as illustrated in Fig. ~\ref{fig:elastic-body}, where the soft tissue varies between 1.0 and 0.5 cm. For brevity, this configuration is hereafter referred to as $\beta_2$. We expect thick areas to exhibit the most significant offset, while slim bodies would exhibit the minimum offset.

A volumetric mesh is generated using the TetGen software \cite{10.1145/2629697}.
The outcome of this process is volumetric meshes, representing the span of $\beta_{2}$ within the SMPL model.
After creating the offset surface, a surface subdivision is performed to increase the tetrahedron density in the volumetric mesh. This step is crucial for FEM simulations to obtain a high-resolution 3D mesh, enabling the generation of detailed simulation results. Surface subdivision and sealing of the edges was performed with Blender ~\cite{Hess:2010:BFE:1893021}.



\subsection{External Forces/ Interactions in Training}
\label{explain_bc_creation}
Interactions with objects are represented as an external force applied to the human body. 
The applied external force can also be parameterized onto a UV map, where each entry $F_i$ corresponds to the magnitude of normal force applied at vertex $V_i$.

Realistic interactions can be with objects of any shape and form. However, we assert that such interactions could be spanned by learning interactions with a small set of simplistic objects.
To put our assertion to the test, we chose to use for training interactions with only circular and elliptical disks, applying constant pressure to the body. This means that the forces used for training are intentionally simplistic compared to the forces applied during the test, which are more complex and diverse in geometry and force magnitude. This distinction is critical as it tests our trained model's ability to generalize from more straightforward learned scenarios to more intricate and varied forces.

\begin{figure}[t]
    \centering
    \begin{subfigure}[t]{0.15\linewidth}
        \centering
        \includegraphics[width=\linewidth]{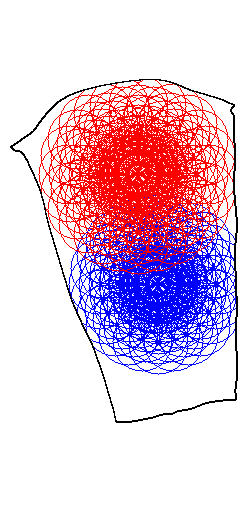}
        \caption{}
        \label{fig:a}
    \end{subfigure}
    \hfill 
    \begin{subfigure}[t]{0.15\linewidth}
        \centering
        \includegraphics[width=\linewidth]{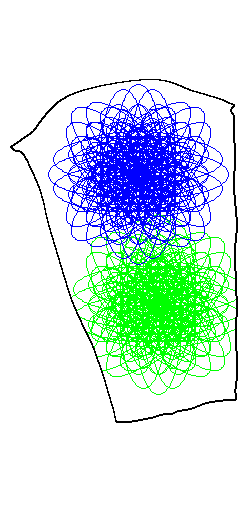}
        \caption{}
        \label{fig:b}
    \end{subfigure}
    \hfill 
    \begin{subfigure}[t]{0.15\linewidth}
        \centering
        \includegraphics[width=\linewidth]{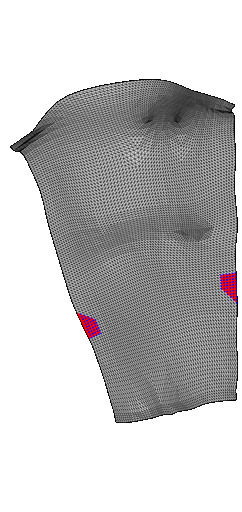}
        \caption{}
        \label{fig:c}
    \end{subfigure}
    \hfill 
    \begin{subfigure}[t]{0.15\linewidth}
        \centering
        \includegraphics[width=\linewidth]{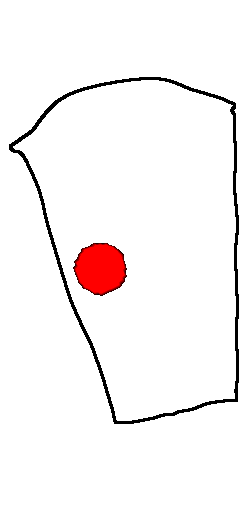}
        \caption{}
        \label{fig:d}
    \end{subfigure}
    \hfill 
    \begin{subfigure}[t]{0.15\linewidth}
        \centering
        \includegraphics[width=\linewidth]{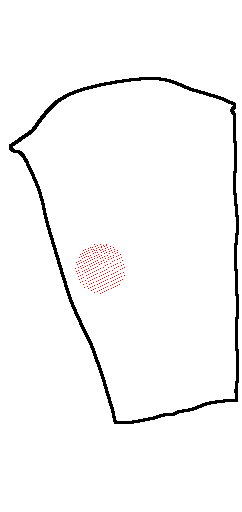}
        \caption{}
        \label{fig:e}
    \end{subfigure}
     \label{fig:BC_5_figures}  
     \caption{
     \textbf{Training Data Span and Examples}
     \ref{fig:a} and \ref{fig:b} illustrate the span of circular and elliptical force area of training data, \ref{fig:c} illustrates an example of seam force area, \ref{fig:d} shows a specific example of interaction with a disk, and \ref{fig:e} shows the vertices activated in simulation to create deformed result. }
\end{figure}

Fig. ~\ref{fig:a} and Fig. ~\ref{fig:b} show the span of circular and elliptical force areas used in training, visualized on top of an arm UV map.
The mapping of the body from 3D into a 2D UV map cuts the 3D shape along a seam. It is thus essential to include interaction with objects across the seam in the training data to avoid inconsistencies, and one such example is illustrated in Fig. ~\ref{fig:c}.
We found that including such data in training is vital for guaranteeing that both sides of the seam get consistent values despite their differing UV coordinates.  

\subsection{Training Data Generation Using FEM Simulation}
\label{training_data_generation}
Ground-truth data for interactions is generated through FEM simulations implementing PolyFEM~\cite{polyfem}. For each positioning of the interacting disks, as described in (Sec. \ref{explain_bc_creation}), calculations are performed with varying normal force strength $F_{total} = [1,2,3,4,5,6,7,8,9,10]$ N. Initially, vertices subjected to a particular boundary condition are identified, with $|V|$ denoting the count of these vertices. Fig. ~\ref{fig:e} exemplifies all vertices under the circular force area shown in Fig. ~\ref{fig:d}. A force boundary condition quantified as:
$F_i = \frac{F_{total}}{|V|}$
is applied to each vertex in the direction normal to the vertex.

\begin{figure}[t]
    \centering
    \includegraphics[width=8cm]{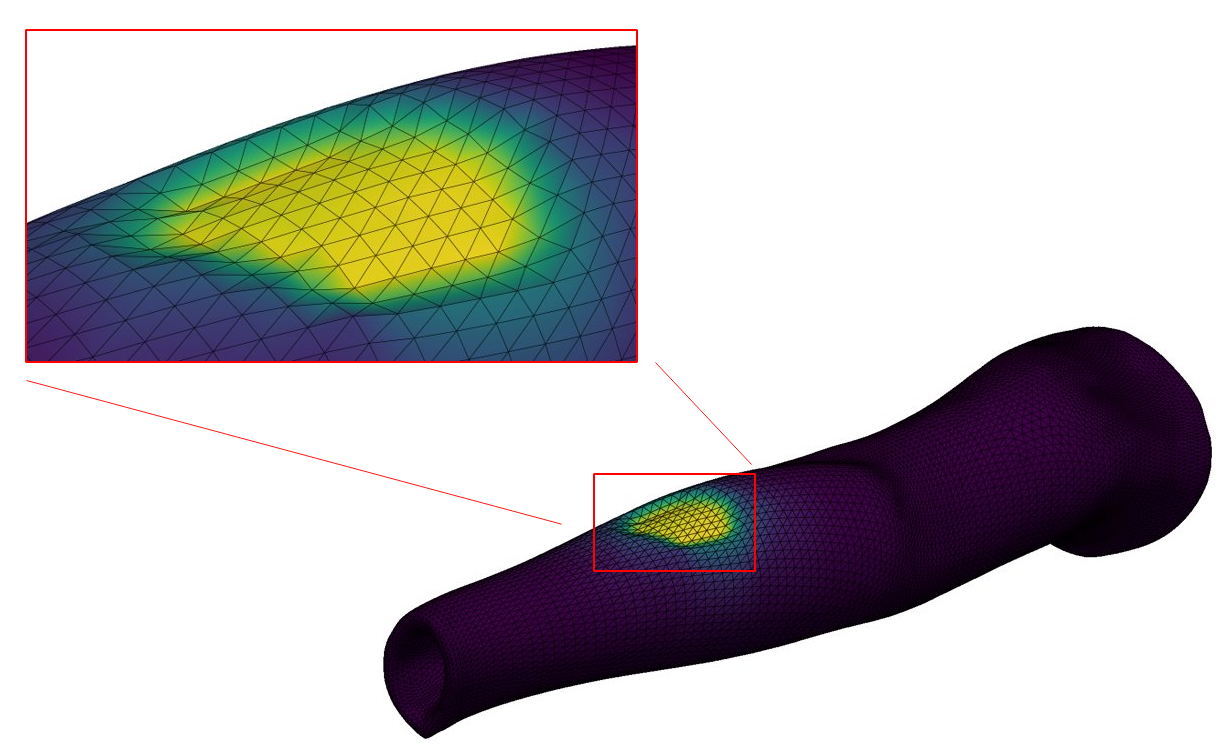}
    \caption{
    \textbf{Example of FEM result}.
    The FEM deformation result under an interaction of force applied on circular area.}
    \label{fig:simulation_result}
\end{figure}

Our dataset consists of a total of 2,655 boundary conditions, including 1,884 circular and 216 elliptical disks and 555 seam conditions. Each boundary condition underwent FEM calculations with various forces, yielding 26,550 simulations per body part, amounting to a total of over 132,000 FEM results. The subsequent data-cleaning process, as explained below, resulted in 97716 viable results. Fig. ~\ref{fig:simulation_result} presents an exemplary FEM outcome for a circular boundary condition.

\textbf{Material Properties}
For the FEM calculation, we need to model the soft tissue's physical properties. Following~\cite{Human_Touch}, the soft tissue of the human body is modeled as hyper-elastic material obeying the Neo-Hookean constitutive law. In the modeling process, Young's modulus is set to $E = 1$ KPa similar to measured values in \cite{molecules24050907}, the Poisson's ratio to $\nu = 0.4$, and the density at $\rho = 1000 \text{ kg/m}^3$.

\textbf{Data Cleaning}
During the FEM calculations, a subset of the data had excessively high local deformations, sometimes displaying unrealistically high values, thereby affecting the overall data integrity. To address this issue, these anomalous examples were filtered out from the dataset. The main strategy for filtering is to limit the maximum applied pressure to 5 KPa, which was achieved empirically by observing that applying greater value pressure resulted in unrealistic results. 

\subsection{From 3D to 2D UV Maps}
\label{3D_to_2D}
To train STMPL, we must translate the body shape, soft tissue, external forces, and FEM-based deformed body onto 2D UV maps.

The UV maps representing the thickness of the soft tissue were normalized to the range [0, 1], with the minimum and maximum values set at [2.5, 12.5] mm.
The UV maps of the external forces were normalized to the range [0, 1], with the minimum and maximum values established at [0, 20] N. 

Each calculated ground-truth deformation is mapped from a 3D volumetric mesh to a 2D UV map. The resultant UV map resembles a pixelized layout where non-zero pixels indicate the deformations of the deformed vertices. These UV maps were then interpolated to create continuous UV images. All deformations were normalized to the range [-1, 1], corresponding to maximum and minimum values of [30, -30] mm.

For each data example, the force UV map with dimensions of [512x512x1] where the pixels represent the normalized magnitude of normal force applied, the deformation UV map with dimensions of [512x512x3] where each pixel represents normalized deformation $[\delta x_i, \delta y_i, \delta z_i]$, and the thickness UV map with dimensions of [512x512x1] where each pixel represents normalized thickness, are concatenated to form a
[512x512x5] tensor. This tensor will later be fed in STMPL as input-output via training. 

\subsection{Implementation Details}
\textbf{Network Architecture.} 
STMPL incorporates UNET architecture, which is distinguished by its fully convolutional network with a U-shaped configuration. 
The UNET architecture we chose is a modified version of~\cite{IsolaZZE16} that aligns with the dimensions and number of channels of our data. We adopt dimensions of 512$\times$512, employing a two-channel input representing force and soft tissue thickness of the simulated arm and a three-channel output representing its deformation when force is applied to it. The UNET operates convolution and transposed convolution with a kernel size of 4, stride 2, dropout 0.5, no normalization, and LeakyReLU with a slope of 0.2 as activation, except the last layer, which uses TanH. We use $L_2$ loss in our training. In some experiments, we tried an addition of Total Variation loss as a smoothness element.

\textbf{Training.} We use Adam optimizer with a learning rate of 1.0 $\times$ $10^{-4}$ for STMPL models and 2.0 $\times$ $10^{-2}$ for Naive model, $L_2$ as a loss function, with a batch size of 64 for STMPL models and 128 for Naive model. Additional details regarding training parameters are provided in the supplementary material. 


\textbf{Inference.}The inference pipeline is illustrated in Figure~\ref{fig:testing_overview}.
We are given as input an interaction between a human body and a complex object, both represented as 3D meshes. During pre-processing, we identify the interaction area between the body and the object and compute the corresponding 2D UV maps of the body shape and the external force. We further require a definition of the body's soft tissue thickness as input, which is also mapped onto the corresponding 2D UV map. 
These UV maps are fed as input into the trained STMPL, which outputs a deformed UV map. 

\textbf{Reconstruction of 3D Deformed Mesh.}
The output deformation UV map generated by the STMPL is remapped back to 3D. This reconstruction process mirrors the initial mapping procedure, wherein deformations were mapped from 3D space to 2D UV map coordinates. Essentially, this inverse operation converts predicted deformations on the UV map into their corresponding 3D forms, ensuring a seamless transition between the two representational states.

\section{Experiments}

\subsection{Dataset and Evaluation Measures}
\label{datasets_description}
\subsubsection{STMPL Dataset.}
A novel dataset, soft tissue deformations of the STMPL model`s left arm, has been created. The choice of the arm stems from the observation that in \cite{santesteban2020softsmpl},\cite{Human_Touch},\cite{Meekyoung:siggraph}, the arms had thinner soft tissue in comparison to other body parts such as the belly, and therefore, resulted in minimal deformation caused by dynamic movement; hence arms was excluded from the simulation.

\begin{figure}[h]
\centering
\begin{minipage}[b]{0.3\linewidth}
    \centering
    \begin{subfigure}[b]{\linewidth}
    \centering
    \includegraphics[width=\linewidth]{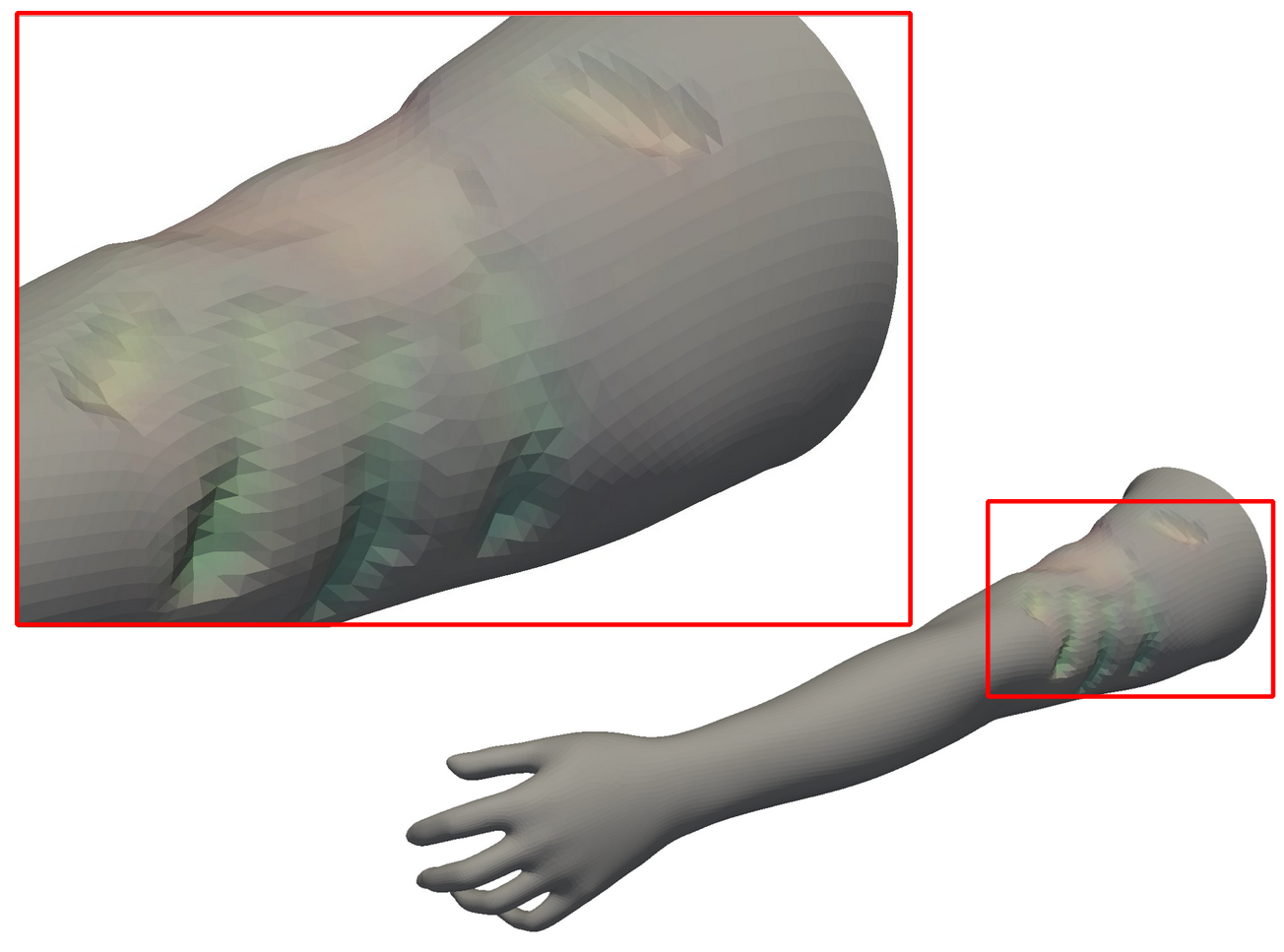}
    \caption{ }
    \label{fig:arm_1}
    \end{subfigure}
\end{minipage}%
\hfill 
\begin{minipage}[b]{0.3\linewidth}
    \centering
    \begin{subfigure}[b]{\linewidth}
    \centering
    \includegraphics[width=\linewidth]{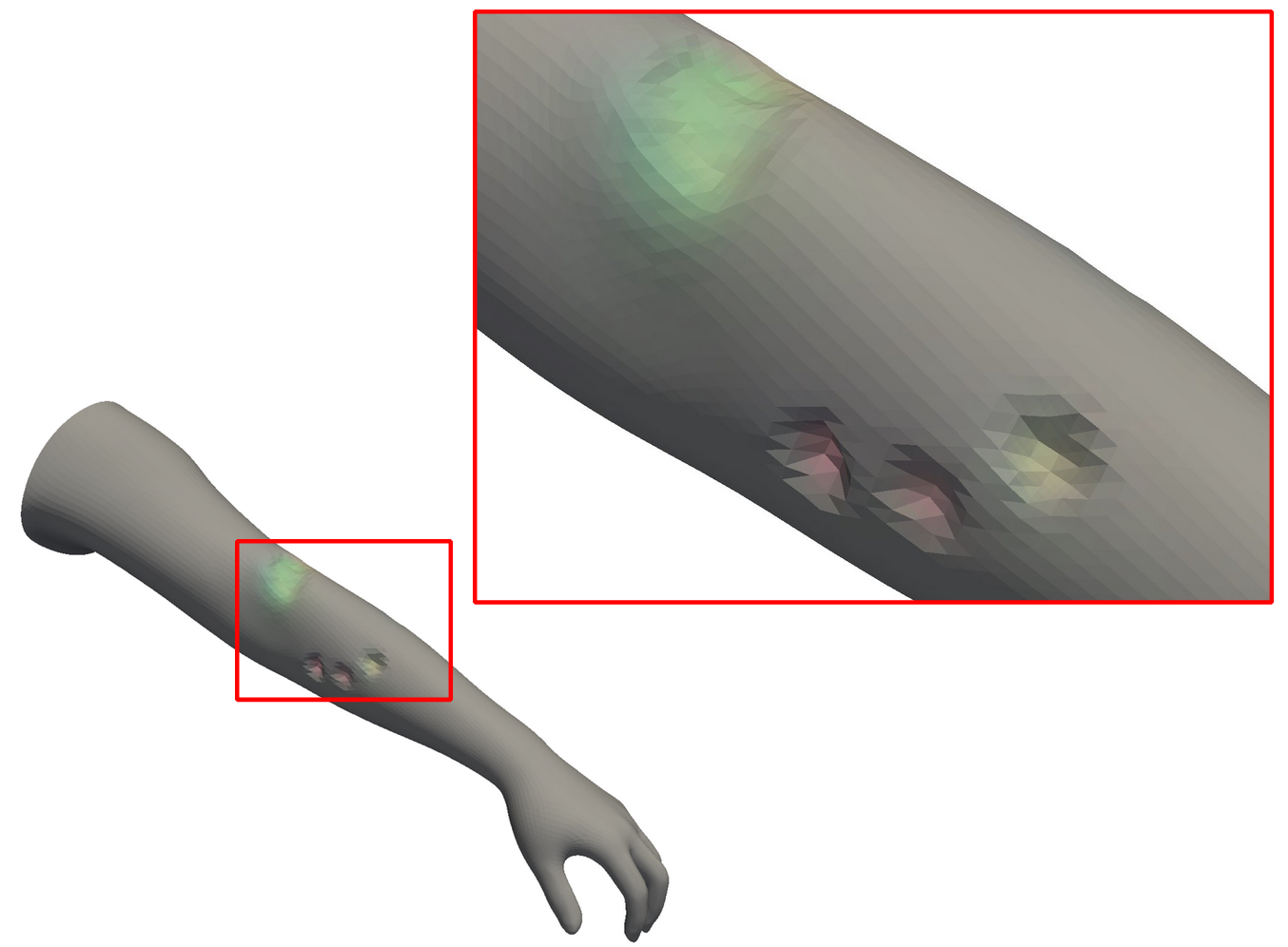}
    \caption{ }
    \label{fig:arm_2}
    \end{subfigure}
\end{minipage}%
\begin{minipage}[b]{0.3\linewidth}
    \centering
    \begin{subfigure}[b]{\linewidth}
    \centering
    \includegraphics[width=\linewidth]{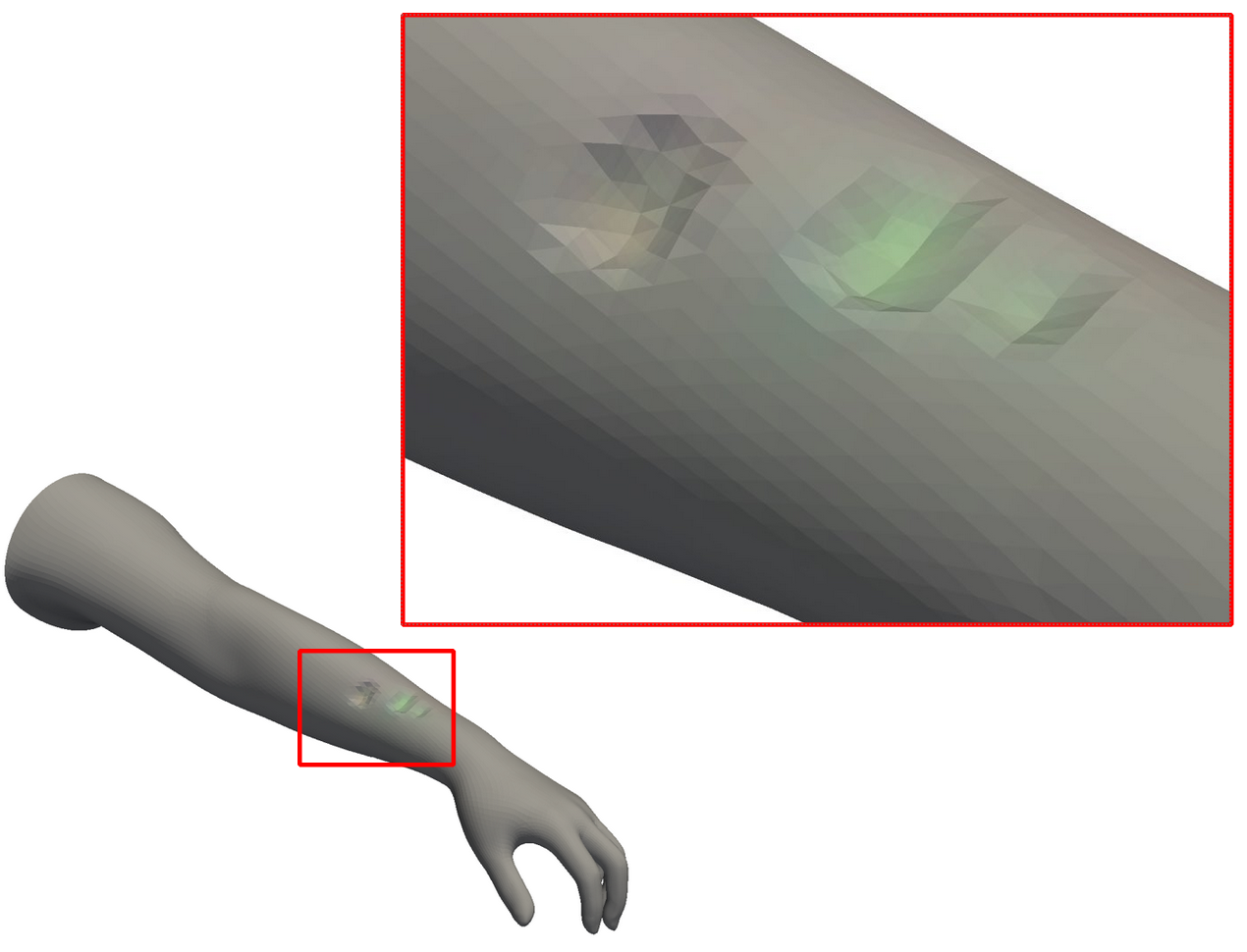}
    \caption{ }
    \label{fig:arm_3}
    \end{subfigure}
\end{minipage}%
\label{fig:test_bc_examples}  
\caption{
\textbf{STMPL Simulated Grasps Examples.}
\ref{fig:arm_1}, \ref{fig:arm_2} and \ref{fig:arm_3} illustrates hand grasping arm examples, which are part of out-of-distribution test set, tested in generalization study.}
\label{fig:test_grabs}  
\end{figure}

\begin{figure}[h]
\centering
\begin{minipage}[b]{0.5\linewidth}
    \centering
    \begin{subfigure}[b]{\linewidth}
        \centering
        \includegraphics[width=\linewidth]{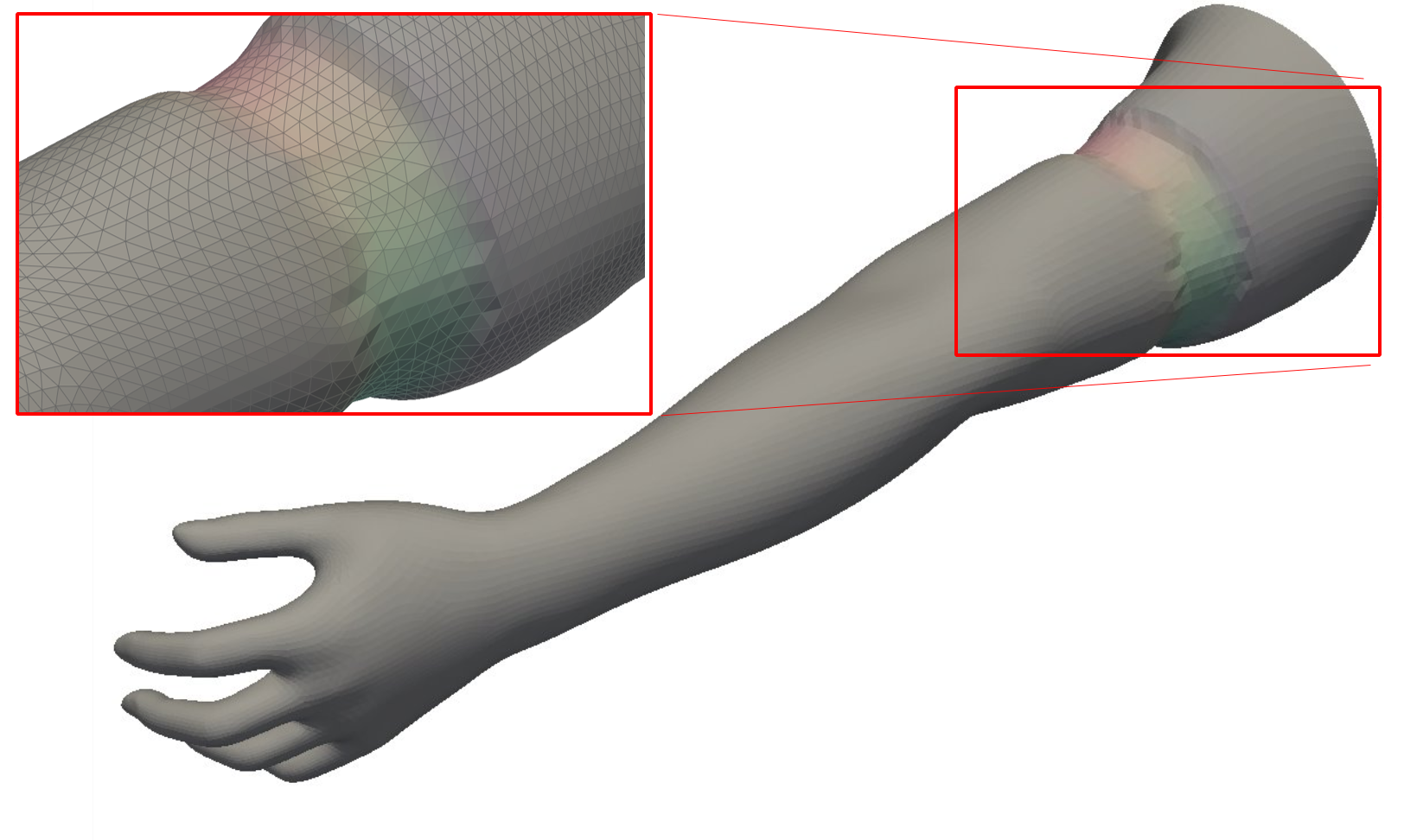}
        \caption{}
        \label{fig:arm_const}
    \end{subfigure}
\end{minipage}%
\hfill 
\begin{minipage}[b]{0.5\linewidth}
    \centering
    \begin{subfigure}[b]{\linewidth}
        \centering
        \includegraphics[width=\linewidth]{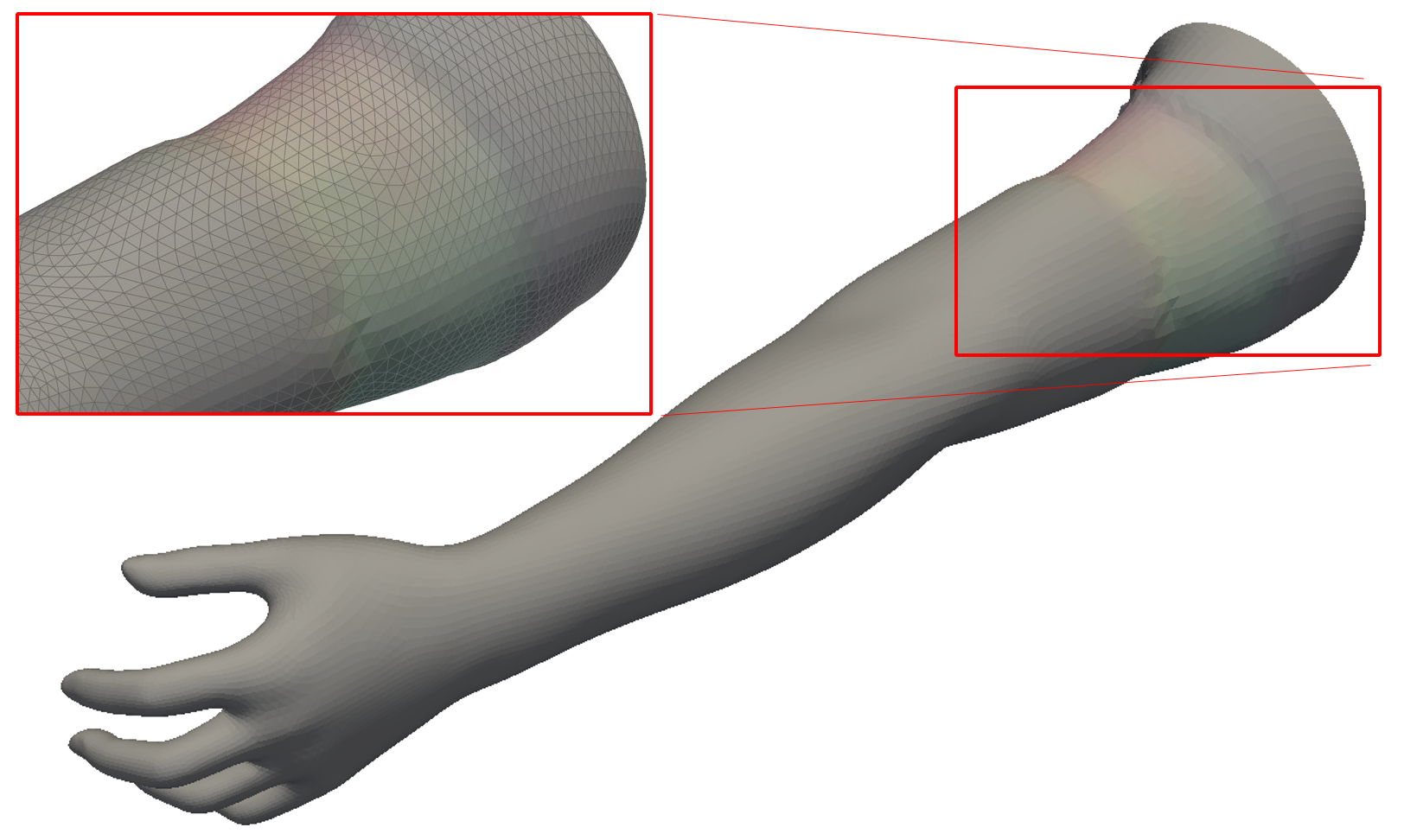}
        \caption{}
        \label{fig:arm_linear}
    \end{subfigure}
\end{minipage}
\caption{\textbf{STMPL Simulates Bands Examples.}
(\subref{fig:arm_const}) and (\subref{fig:arm_linear}) illustrate bands pressing on arm examples, which are part of the out-of-distribution test set, tested in a generalization study.}
\label{fig:test_bands}  
\end{figure}

\begin{figure}[!]
\centering
\begin{minipage}[b]{0.3\linewidth}
    \centering
    \begin{subfigure}[b]{\linewidth}
        \centering
        \includegraphics[width=\linewidth]{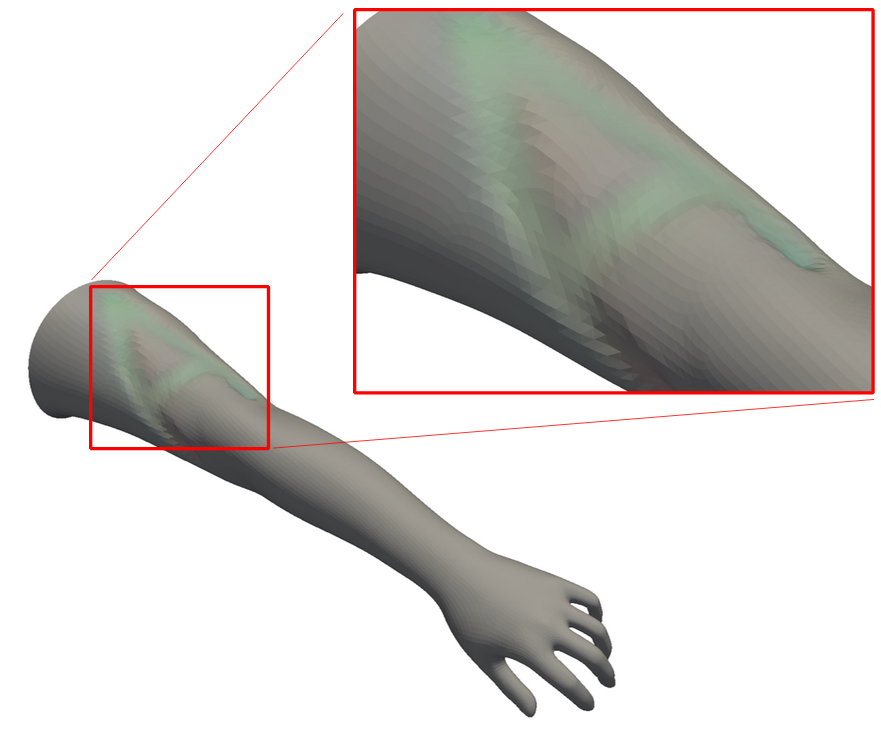}
        \caption{}
        \label{fig:arm_A}
    \end{subfigure}
\end{minipage}%
\hfill 
\begin{minipage}[b]{0.3\linewidth}
    \centering
    \begin{subfigure}[b]{\linewidth}
        \centering
        \includegraphics[width=\linewidth]{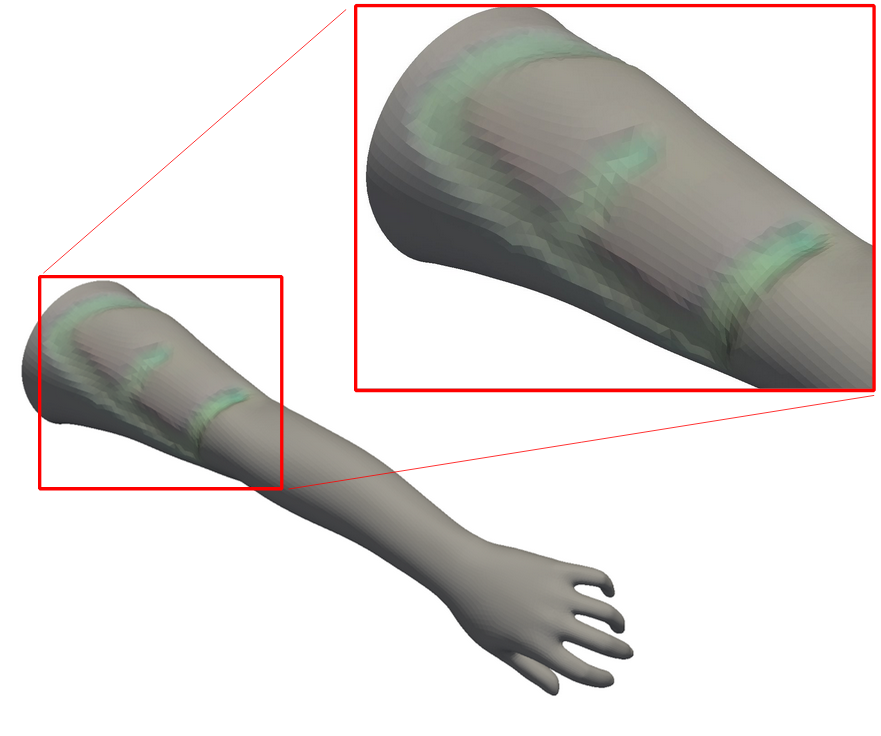}
        \caption{}
        \label{fig:arm_E}
    \end{subfigure}
\end{minipage}%
\begin{minipage}[b]{0.3\linewidth}
    \centering
    \begin{subfigure}[b]{\linewidth}
        \centering
        \includegraphics[width=\linewidth]{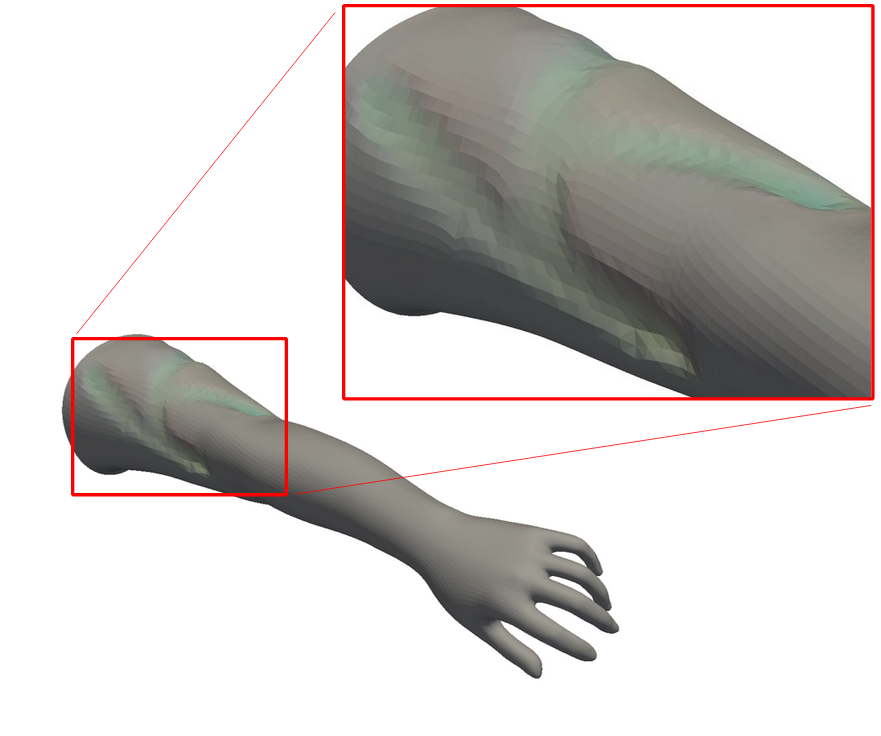}
        \caption{}
        \label{fig:arm_K}
    \end{subfigure}
\end{minipage}%
     \label{fig:test_bc_examples_letters}
     \caption{\textbf{STMPL Simulates Letters Examples.}
     \ref{fig:arm_A}, \ref{fig:arm_E} and \ref{fig:arm_K} illustrate block letters A, E, and K force applied on the arm, which are part of an out-of-distribution test set, tested in generalization study.}
\label{fig:test_letters} 
\end{figure}

In terms of body shapes and soft tissue thickness, our dataset covers the domain of $\beta_{2}$, directly influencing the quantity of soft tissue in the human body. A deliberate selection was made to encompass both extremities of the $\beta_{2}$ range, with values set at 2.0 and -2.0 and three intermediate values: 1.0, 0.0, and -1.0. 
Our dataset comprises 97716 deformations of simple interactions such as circular and elliptical discs described in Sec. \ref{explain_bc_creation} encompassing all five $\beta_{2}$.
This dataset was split into training, validation, and testing with a 0.8/0.1/0.1 ratio. We call the test split of this dataset an in-distribution test set. Furthermore, we created a second dataset, which we call the out-of-distribution test set, which consists of three sets of testing dedicated data. This dataset was created by applying interactions with various complex shapes to thoroughly evaluate the model's performance under diverse and intricate scenarios such as hand grasps (395 examples) shown in Fig.\ref{fig:test_grabs}, bands on the arm 40 examples as shown in Fig.\ref{fig:test_bands}, and alphabet letters contour force application (273 examples), shown in Fig.\ref{fig:test_letters}.

\subsubsection{Training Setups.}
\label{training_setups}
We evaluate the performance of three training setups:

\begin{enumerate}[(i)]
    \item STMPL model: trained on the full train set.
    \item STMPL model only 3 $\beta_2$: same model as (i), but trained on the part of train set which spans $\beta_2 = [2.0,0.0,-2.0]$.
    \item Naive model: a baseline method, defined in \ref{baseline_model_section}, trained on the full train set.
\end{enumerate}

\subsubsection{Evaluation Measures.}
Each test sample underwent evaluation by identifying ground-truth vertices with deformation greater than $0.001$ mm, which then served as focal points for calculating performance measures. The Mean Absolute Error (MAE) in the x, y, and z directions within global coordinates is computed and subsequently reported as MAE x, MAE y, and MAE z. The Mean Euclidean Error (MEE) was also calculated and reported accordingly. Additionally, we show the ratio of MEE under [0.1,0.125,0.15,0.2,0.5] mm following \cite{PhysGNN}.

\subsection{Baseline Model Creation}
\label{baseline_model_section}
We establish a simple baseline method for comparison, termed the Naive model. This model predicts deformation using the following definition:

\begin{equation}\label{Naive_eq}
\begin{aligned}
\delta x &= \alpha_x \cdot t_i \cdot F_i \\
\delta y &= \alpha_y \cdot t_i \cdot F_i \\
\delta z &= \alpha_z \cdot t_i \cdot F_i
\end{aligned}
\end{equation}

where $t_i$ is the thickness per vertex and $F_i$ is force per vertex and $\alpha_x$, $\alpha_y$ and $\alpha_z$ are learnable parameters.

\subsection{Evaluation}
\subsubsection{Test Results.}
To evaluate the performance of trained models described in \ref{training_setups}, we first test them on the in-distribution test set, described in \ref{datasets_description}, similar to training data. Table~\ref{tab:table_1} shows that in both STMPL models, at least 91\% errors are under 0.1 mm, whereas in the Naive model, the error is under 0.1 mm in 46.82\% of test cases.

 Next, we evaluate the performance of all trained models testing with an out-of-distribution test set. Table~\ref{tab:table_2} shows that between STMPL models, the best performing, as expected, is trained on all training data and archives error under 0.1 mm in 49.3\% of test scenarios, whereas the Naive model performs with error under 0.1 mm in 37.9\% of test scenarios. As shown in Table \ref{tab:table_2}, MEE for STMPL, which trained on all training data is $0.116\pm0.0666$ mm where maximum deformation is $2.081\pm1.7295$ mm on the test data is, which makes relative error $5.57\%$. As anticipated, the models exhibit decreased performance when evaluated on out-of-distribution data compared to in-distribution data.
 
\begin{figure}[!]
    \centering
    \includegraphics[width=12cm]{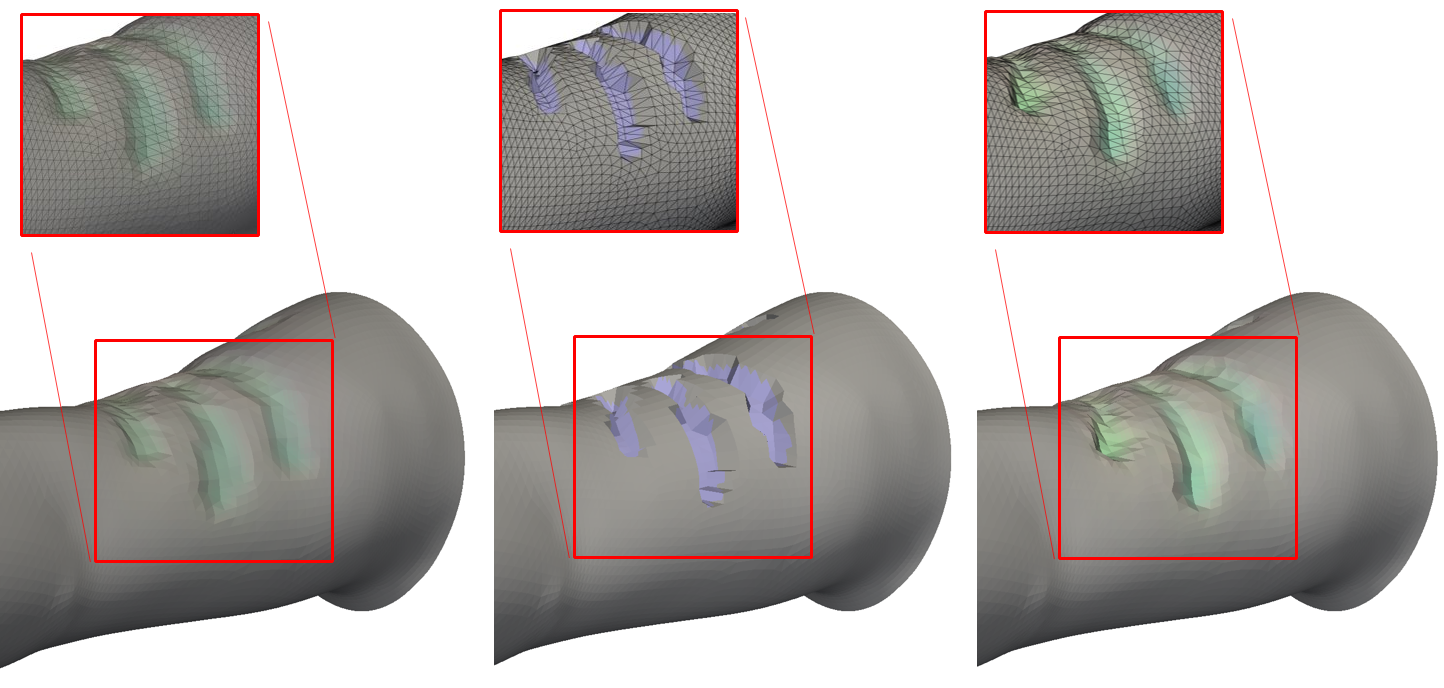}
    \caption{
    \textbf{Qualitative Comparison of FEM with Naive Baseline and STMPL (ours)}.
    We visually compare simulated results of STMPL (right) of the hand grasp scenario (out-of-distribution example) with FEM ground truth (left) and Naive baseline model (middle).
    }
    \label{fig:gt_naive_stmpl_comparison}
\end{figure}

 Fig. \ref{fig:gt_naive_stmpl_comparison} illustrates a visual comparison between FEM ground truth (left), Naive model (middle), and STMPL model (right) simulated result of hand grasping arm with a total applied force of 20 N. This Figure clearly shows the Naive model constant deformation and the deformation of STMPL is qualitatively similar to FEM ground truth.

\begin{table}[!t]
\fontsize{7pt}{8pt}
\selectfont
\caption
{
A quantitative comparison of the STMPL model with the baseline "Naive" model, tested on an in-distribution test set. Mean Absolute Error (MAE), Mean Euclidean Error (MEE), and the percentage of predictions with Euclidean Error (EE) below 0.1, 0.125, 0.15, 0.2, 0.5 mm.
}
\label{tab:table_1}
\centering
\begin{tabular}{|l|c|c|c|c|c|c|c|c|c|c }
\hline

\Longstack{Experiment Name} & \Longstack{MAE \\ $\delta x$ \\ (mm) }   & \Longstack{MAE \\ $\delta y$ \\ (mm) } & \Longstack{MAE \\ $\delta z$ \\ (mm) } & \Longstack{MEE \\ \\ (mm) } & \Longstack{\\ EE $\leq$ \\ 0.1 \\ mm \\ (\%)} & \Longstack{\\ EE $\leq$ \\ 0.125 \\ mm \\ (\%)} & \Longstack{\\ EE $\leq$ \\ 0.15 \\ mm \\ (\%)} & \Longstack{\\ EE $\leq$ \\ 0.2 \\ mm \\ (\%)} & \Longstack{\\ EE $\leq$ \\ 0.5 \\ mm \\ (\%)} \\
\hline

STMPL &
  $0.0196\pm$ $ 0.0098$ &
  $0.0321\pm$ $ 0.0251$ &
  $0.0339\pm$ $ 0.0269$ &
  $0.0577\pm$ $ 0.0408$ &
  91.13 &
  94.98 &
  97.08 &
  98.58 &
  99.94 \\
\textbf{STMPL: only 3 $ \beta_2 $}  &
 \textbf{0.0182$\pm$ 0.0098} &
  \textbf{0.0301$\pm$ 0.0262} &
  \textbf{0.0316$\pm$  0.0281} &
  \textbf{0.0545$\pm$  0.0423} &
  \textbf{91.18} &
  \textbf{94.9} &
  \textbf{96.85} &
  \textbf{98.76} &
  \textbf{99.95} \\
Naive &
  $0.0347\pm$ $ 0.0148$ &
  $0.0637\pm$ $ 0.0338$ &
  $0.0632\pm$ $ 0.0336$ &
  $0.1106\pm$ $ 0.0505$ &
  46.82 &
  70.21 &
  83.9 &
  94.59 &
  99.95 \\

\hline
\end{tabular}
\end{table}
\begin{table}[t]
\fontsize{7pt}{8pt}
\selectfont
\caption{
A quantitative comparison of the STMPL model with the baseline "Naive" model, tested on an out-of-distribution test set. Mean Absolute Error (MAE), Mean Euclidean Error (MEE), and the percentage of predictions with Euclidean Error (EE) below 0.1, 0.125, 0.15, 0.2, 0.5 mm.}
\label{tab:table_2}
\centering
\begin{tabular}{|l|c|c|c|c|c|c|c|c|c|c }
\hline

\Longstack{Experiment Name} & \Longstack{MAE \\ $\delta x$ \\ (mm) }   & \Longstack{MAE \\ $\delta y$ \\ (mm) } & \Longstack{MAE \\ $\delta z$ \\ (mm) } & \Longstack{MEE \\ \\ (mm) } & \Longstack{\\ EE $\leq$ \\ 0.1 \\ mm \\ (\%)} & \Longstack{\\ EE $\leq$ \\ 0.125 \\ mm \\ (\%)} & \Longstack{\\ EE $\leq$ \\ 0.15 \\ mm \\ (\%)} & \Longstack{\\ EE $\leq$ \\ 0.2 \\ mm \\ (\%)} & \Longstack{\\ EE $\leq$ \\ 0.5 \\ mm \\ (\%)} \\
\hline

\textbf{STMPL} &
  \textbf{0.0346$\pm$  0.0208} &
  \textbf{0.0648$\pm$  0.0411} &
  \textbf{0.0695$\pm$  0.0390} &
  \textbf{0.1160$\pm$  0.0666} &
  \textbf{49.3} &
  \textbf{63.92} &
  \textbf{74.38} &
  \textbf{88.17} &
  \textbf{99.94} \\
STMPL: only 3 $ \beta_2 $  &
  $0.0345\pm$ $ 0.0147$ &
  $0.0719\pm$ $ 0.0398$ &
  $0.0801\pm$ $ 0.0439$ &
  $0.1289\pm$ $ 0.0662$ &
  36.64 &
  56.16 &
  70.77 &
  86.22 &
  100.0 \\
Naive &
  $0.0444\pm$ $ 0.0303$ &
  $0.1015\pm$ $ 0.0889$ &
  $0.1095\pm$ $ 0.0993$ &
  $0.1775\pm$ $ 0.1527$ &
  37.9 &
  48.37 &
  57.71 &
  69.85 &
  96.11 \\

\hline
\end{tabular}
\end{table}
\begin{table}[t]
\fontsize{7pt}{8pt}
\selectfont
\caption{\textbf{Generalization Ablation Test}. STMPL: only 3 $\beta_2$ model tested on different sets of data to evaluate its generalization.}
\label{tab:table_3}
\centering
\begin{tabular}{|l|c|c|c|c|c|c|c|c|c|c }
\hline

\Longstack{ Experiment Name } & \Longstack{MAE \\ $\delta x$ \\ (mm) }   & \Longstack{MAE \\ $\delta y$ \\ (mm) } & \Longstack{MAE \\ $\delta z$ \\ (mm) } & \Longstack{MEE \\ \\ (mm) } & \Longstack{\\ EE $\leq$ \\ 0.1 \\ mm \\ (\%)} & \Longstack{\\ EE $\leq$ \\ 0.125 \\ mm \\ (\%)} & \Longstack{\\ EE $\leq$ \\ 0.15 \\ mm \\ (\%)} & \Longstack{\\ EE $\leq$ \\ 0.2 \\ mm \\ (\%)} & \Longstack{\\ EE $\leq$ \\ 0.5 \\ mm \\ (\%)} \\
\hline

\begin{tabular}[c]{@{}l@{}}In-distribution\\ seen thickness\end{tabular}  &
  $0.0189\pm$ $ 0.01$ &
  $0.0313\pm$ $ 0.0259$ &
  $0.0332\pm$ $ 0.0289$ &
  $0.0569\pm$ $ 0.0425$ &
  90.25 &
  94.44 &
  96.63 &
  98.56 &
  99.96 \\
\hline
\begin{tabular}[c]{@{}l@{}}In-distribution\\ unseen thickness\end{tabular}  &
  $0.0172\pm$ $ 0.0093$ &
  $0.0287\pm$ $ 0.0265$ &
  $0.0296\pm$ $ 0.0268$ &
  $0.0515\pm$ $ 0.0418$ &
  92.37 &
  95.48 &
  97.13 &
  99.0 &
  99.93 \\
\hline
\begin{tabular}[c]{@{}l@{}}Out-of-distribution\\ seen thickness\end{tabular}  &
  $0.0368\pm$ $ 0.0159$ &
  $0.0771\pm$ $ 0.0431$ &
  $0.0845\pm$ $ 0.0462$ &
  $0.137\pm$ $ 0.0707$ &
  33.18 &
  50.77 &
  65.56 &
  82.58 &
  100.0 \\
\hline
\begin{tabular}[c]{@{}l@{}}Out-of-distribution\\ unseen thickness\end{tabular}  &
  $0.0282\pm$ $ 0.0079$ &
  $0.0578\pm$ $ 0.0236$ &
  $0.0678\pm$ $ 0.034$ &
  $0.1066\pm$ $ 0.0448$ &
  46.15 &
  70.96 &
  85.09 &
  96.23 &
  100.0 \\
\hline
\end{tabular}
\end{table}

\subsection{Generalization Test}
To explore the generalization abilities of our approach, we tested the performance of the STMPL: only 3 $\beta_2$, which was trained on the part of training data, which consists only of $\beta$ = [2.0,0.0,-2.0], as described in \ref{training_setups}. We test this model's performance on four parts of previously defined test sets, two from in-distribution and two from out-of-distribution sets.
The first two are test data from an in-distribution test set, where the first consists of $\beta$ = [2.0,0.0,-2.0], which we call "in-distribution seen thickness," and the second consists of $\beta$ = [1.0,-1.0], which we call "in-distribution unseen thickness.". The second two are test data from an out-of-distribution test set, where the first consists of $\beta$ = [2.0,0.0,-2.0], which we call "out-of-distribution seen thickness," and the second consists of $\beta$ = [1.0,-1.0], which we call "out-of-distribution unseen thickness."
In Table~\ref{tab:table_3}, we demonstrate that our model exhibits generalization capabilities when applied to unseen thicknesses and test examples that are out-of-distribution with train examples. Specifically, the model trained on $\beta_2$ values of [2.0, 0.0, -2.0] shows robust performance when tested on a range of $\beta_2$ values including [2.0, 1.0, 0.0, -1.0, -2.0]. The MEE for predictions on data that are "out-of-distribution unseen thickness" in the training data is $0.1066\pm 0.0448$ mm, which makes this result relative error to max deformation of the test set $5.32\%$, and for "out-of-distribution seen thickness" is $0.137\pm 0.0707$ mm, which we hypothesize larger because on average the deformation itself in [2.0, 0.0, -2.0] is larger than in [1.0,-1.0].

\subsection{Runtime Evaluation}
According to the result from Table~\ref{tab:table_2}, STMPL can predict left arm soft tissue deformation under various loads on out-of-distribution data with the error of 0.116 $\pm$ 0.0666 mm. Approximating those deformations using the FEM on Intel i7-12700K CPU took $84\pm53.4$  seconds on average, simulating on one thread, while on NVIDIA A6000 GPU took $0.002\pm0.0003$ seconds on average.

\section{Conclusion}
This paper introduced STMPL, a data-driven soft-tissue simulator of human body parts. Our method augments SMPL with additional soft layer representation and simulates the deformation of this soft layer under contact interaction. Our experiments indicate that our model shows good generalization to an unseen arm thickness and also on test data that is out-of-distribution to train data.

\textbf{Limitations and Future Work.}
Although we only report results for arms with constant thickness, our method can be extended beyond this case.
The most exciting direction for future work is to extend this work to modeling the entire human body.
While our method has been shown effective in simulating soft-tissue deformation in human body parts in our setting, it is a possible future direction to model body parts with non-constant thickness and variable material properties. An additional interesting future direction is to try a different approach to a data-driven simulator based on Graph Neural Networks. A possible approach in this direction is to train a small volume patch with different material properties and, in inference time, simulate in a specific area, extending a pre-trained small patch to achieve full-area simulation.

%
%
\bibliographystyle{splncs04}
\bibliography{main}
\end{document}